\documentclass[letterpaper]{article} % DO NOT CHANGE THIS
\usepackage{aaai24}  % DO NOT CHANGE THIS
\usepackage{times}  % DO NOT CHANGE THIS
\usepackage{helvet}  % DO NOT CHANGE THIS
\usepackage{courier}  % DO NOT CHANGE THIS
\usepackage[hyphens]{url}  % DO NOT CHANGE THIS
\usepackage{graphicx} % DO NOT CHANGE THIS
\usepackage{natbib}  % DO NOT CHANGE THIS AND DO NOT ADD ANY OPTIONS TO IT
\usepackage{caption} % DO NOT CHANGE THIS AND DO NOT ADD ANY OPTIONS TO IT
\usepackage{algorithm}
\usepackage{algorithmic}

\usepackage{multirow}
\usepackage{amsmath}
\usepackage{caption}
\usepackage{amssymb}
\newcommand{\ie}{\textit{i}.\textit{e}.}
\newcommand{\eg}{\textit{e}.\textit{g}.}

\usepackage{newfloat}
\usepackage{listings}
\DeclareCaptionStyle{ruled}{labelfont=normalfont,labelsep=colon,strut=off} % DO NOT CHANGE THIS
\floatstyle{ruled}
\newfloat{listing}{tb}{lst}{}
\floatname{listing}{Listing}
\title{Decoupling Degradations with Recurrent Network for Video Restoration in Under-Display Camera}

\author{
    Chengxu Liu\textsuperscript{\rm 1,2}, 
    Xuan Wang\textsuperscript{\rm 3}, 
    Yuanting Fan\textsuperscript{\rm 1}, 
    Shuai Li\textsuperscript{\rm 3}, 
    Xueming Qian\textsuperscript{\rm 1,2} 
}
\affiliations{
    \textsuperscript{\rm 1}Xi’an Jiaotong University\\
    \textsuperscript{\rm 2}Shaanxi Yulan Jiuzhou Intelligent Optoelectronic Technology Co., Ltd\\
    \textsuperscript{\rm 3}MEGVII Technology\\
    liuchx97@gmail.com, retofan@stu.xjtu.edu.cn, \{wangxuan02,lishuai\}\noindent@megvii.com, qianxm@mail.xjtu.edu.cn
}

\usepackage{bibentry}

\begin{document}

\maketitle

\begin{abstract}
Under-display camera (UDC) systems are the foundation of full-screen display devices in which the lens mounts under the display. The pixel array of light-emitting diodes used for display diffracts and attenuates incident light, causing various degradations as the light intensity changes. Unlike general video restoration which recovers video by treating different degradation factors equally, video restoration for UDC systems is more challenging that concerns removing diverse degradation over time while preserving temporal consistency. In this paper, we introduce a novel video restoration network, called D$^2$RNet, specifically designed for UDC systems. It employs a set of Decoupling Attention Modules (DAM) that effectively separate the various video degradation factors. More specifically, a soft mask generation function is proposed to formulate each frame into flare and haze based on the diffraction arising from incident light of different intensities, followed by the proposed flare and haze removal components that leverage long- and short-term feature learning to handle the respective degradations. Such a design offers an targeted and effective solution to eliminating various types of degradation in UDC systems. We further extend our design into multi-scale to overcome the scale-changing of degradation that often occur in long-range videos. To demonstrate the superiority of D$^2$RNet, we propose a large-scale UDC video benchmark by gathering HDR videos and generating realistically degraded videos using the point spread function measured by a commercial UDC system. Extensive quantitative and qualitative evaluations demonstrate the superiority of D$^2$RNet compared to other state-of-the-art video restoration and UDC image restoration methods. Code is available at \url{https://github.com/ChengxuLiu/DDRNet.git}.
\end{abstract}

\section{Introduction}
\label{sec:1}
The rising popularity of full-screen mobile devices has driven the development of under-display camera (UDC) systems. While research on UDC primarily focuses on single image restoration~\cite{feng2022mipi,zhou2020udc}, few works are available on video restoration, which impedes the popularity of devices with UDC systems.
UDC system is an imaging system where the lens is mounted under the display. It can eliminate the screen notch of the traditional front camera in mobile devices, providing a bezel-less display without disrupting the screen’s integrity~\cite{qin2021p}. 

\begin{figure}[t]
  \begin{center}
  \includegraphics[width=1.0\linewidth,page=1]{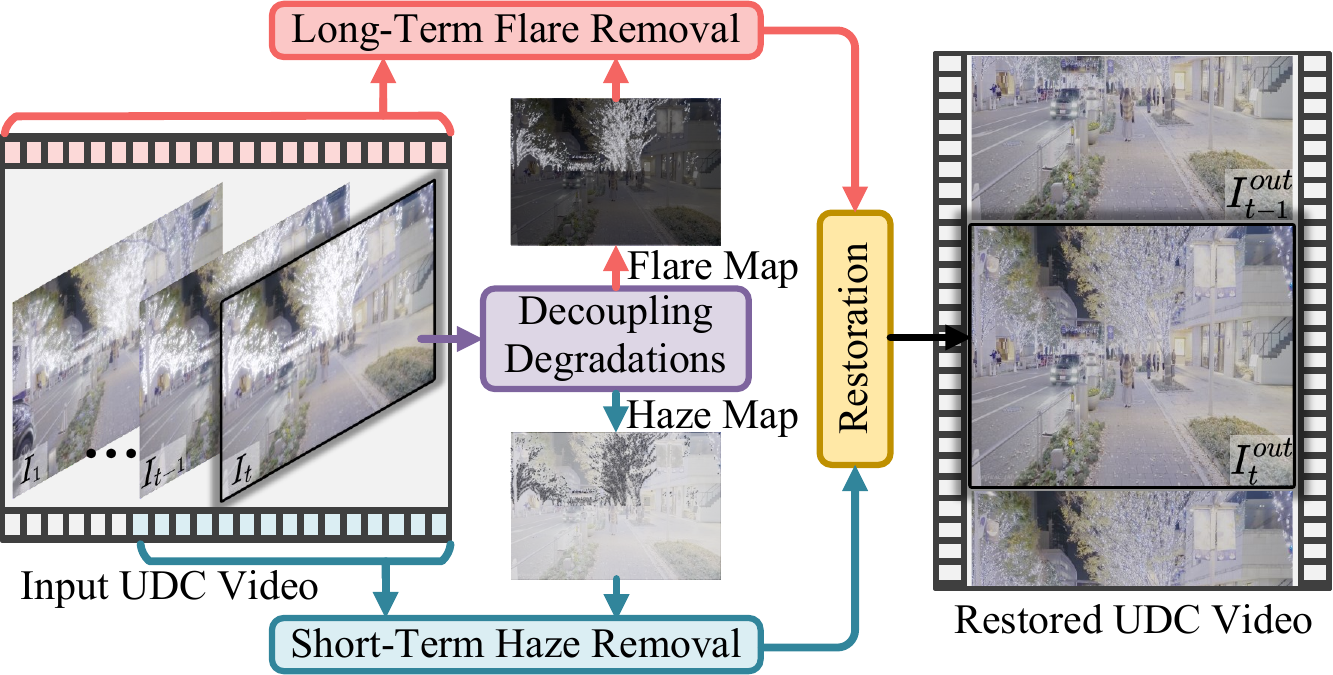}
  \end{center}
  \caption{Method illustration. In UDC systems, the degree of degradation is positively correlated with the intensity of incident light. Our method decouples the degradation into brighter flare and darker haze, which are handled using information from long and short distances, respectively.}
  \label{fig:teaser}
\end{figure}

\begin{figure*}[t]
  \begin{center}
  \includegraphics[width=0.95\linewidth,page=1]{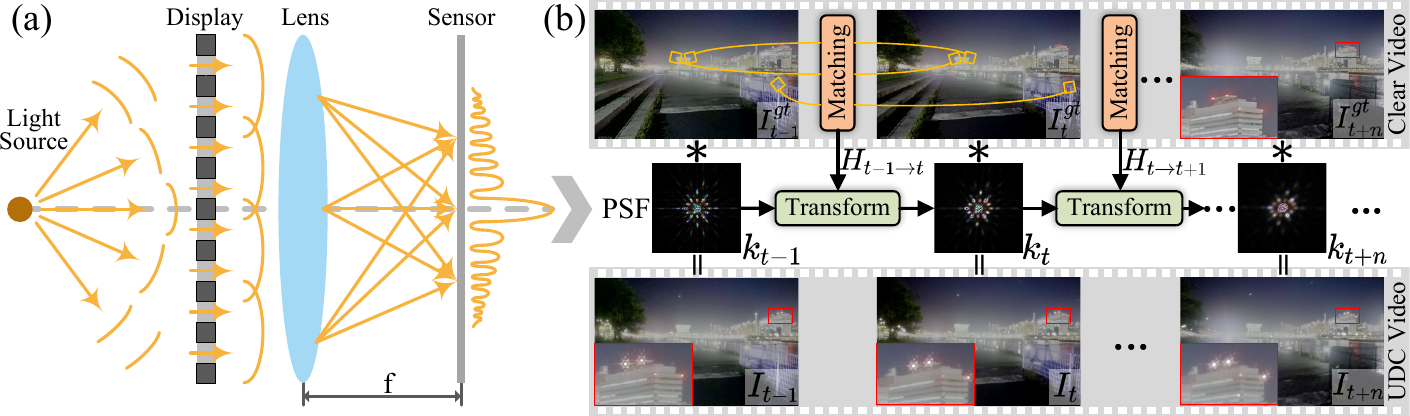}
  \end{center}
  \caption{(a) illustrates the formation of the PSF in UDC systems. The light emitted from the light source crosses a display and a lens before it is finally captured by the sensor. (b) is the generation of UDC video, where the matching part computes the Homography matrix (\ie,~$H$) corresponding to inter-frame motion, and the transform part performs perspective warp on PSF.}
\label{fig:image}
\end{figure*}

In contrast to conventional cameras, during UDC imaging, the incident light will cross the densely arranged organic light-emitting diodes (OLEDs) used for display before arriving at the lens. It implies that incident light is diffracted when propagating the aperture between the OLEDs, especially when the wavelength of the light is similar to the gaps between the obstacles~\cite{zhou2021image} (illustrated by Fig.~\ref{fig:image}(a)). Besides, the degree of degradation arising from diffraction is positively correlated with the intensity of incident light~\cite{kwon2021controllable}. As depicted in Fig.~\ref{fig:teaser}, in brighter regions close to the light source, diffraction causes \textbf{flare} that saturates one or more channels of the image, resulting in content loss. In contrast, in other darker regions, diffraction causes \textbf{haze} that makes the content fuzzy.

To solve these challenges, many efforts have been devoted to handle image restoration for UDC through deep learning-based models. These works can be categorized into two paradigms. Some attempts to leverage the prior knowledge of the diffraction blur kernel, \ie,~point spread function (PSF) illustrated in Fig.~\ref{fig:image}(b), to guide the removal of diffraction~\cite{feng2021removing,kwon2021controllable,liu2022udc}. Another part directly learns diffraction removal through complex network design~\cite{feng2023generating,koh2022bnudc,liu2023fsi}. Unfortunately, unlike UDC images, diffraction will change dynamically with motion in UDC video. Therefore, existing methods for images are unable to take advantage of the strong temporal coherence of diffraction over time, leading to poor performance.

From a methodology perspective, unlike image restoration which only learns on spatial dimensions, video restoration pays more attention to exploiting temporal information. Existing video restoration methods either align features of adjacent frames (\eg,~5 or 7 frames) through a sliding window input mechanism~\cite{liang2022vrt}, or learn features from the more distant frame through a recurrent mechanism~\cite{liang2022recurrent}. Among them, benefiting from the long-term modeling capability of recurrent mechanisms, significant progress has been made in video super-resolution~\cite{liu2022learning}, deblurring~\cite{zhong2023real}, and denoise~\cite{tassano2020fastdvdnet} tasks. 
For UDC videos with complex degradations, the distant frames have more content differences but help recover lost content through the recurrent mechanism~\cite{liu2022learning,chan2022basicvsrpp,liu2022ttvfi} (\ie,~eliminate flare). Exploiting more similar scene patterns from adjacent frames in spatio-temporal neighborhood is essential for recovering clear content~\cite{wang2022efficient,lin2022flow,zhang2022spatio} (\ie,~eliminate haze). Therefore, a more promising solution is to explore proper network with long- and short-term video representation learning to effectively and pertinently eliminate various degradations in UDC video.

In this paper, we propose a novel UDC video restoration network to enable effective video representation learning (D$^2$RNet). The key idea of D$^2$RNet is to decouple the degradations in UDC videos while recovering them separately with different features pertinently (as shown in Fig~\ref{fig:teaser}). To achieve this, we propose a decoupling attention module (DAM) in conjunction with a globally multi-scale bi-directional recurrent framework. In particular, a soft mask generation function is used to partition each frame into flare and haze regions, which are produced by the diffraction of strongly and weakly incident light, respectively. For the flare region, a flare removal component learns long-term features to recover the content loss. For the haze region, a haze removal component learns short-term features to recover content fuzzy. DAM is extended to three scales to overcome the scale-changing of degradation in long-range UDC videos. Besides, for evaluation, we establish a large-scale UDC video restoration benchmark, dubbed VidUDC33K. It contains 677 paired videos of length 50 with 1080p resolution, covering various challenging scenarios.

Our contributions are summarized as follows:
\begin{itemize}
\item We propose a novel network with long- and short-term video representation learning by decoupling video degradations for the UDC video restoration task (D$^2$RNet), which is the first work to address UDC video degradation. The core decoupling attention module (DAM) enables a tailored solution to the degradation caused by different incident light intensities in the video. 
\item We propose a large-scale UDC video restoration dataset (VidUDC33K), which includes numerous challenging scenarios. To the best of our knowledge, this is the first dataset for UDC video restoration.
\item Extensive quantitative and qualitative evaluations demonstrate the superiority of D$^2$RNet. In the proposed VidUDC33K dataset, D$^2$RNet gains 1.02db PSNR improvements more than other restoration methods.
\end{itemize}

\section{Related Work}

\subsubsection{UDC Restoration.}
Recently, UDC restoration based on deep learning has made significant progress and become an increasingly promising research topic. Existing benchmarks are mainly studies based on images. Typically, MSUNet~\cite{zhou2021image} proposes a first UDC image restoration benchmark by analyzing the optical imaging process of real UDC for ECCV20 challenge~\cite{zhou2020udc}. It includes diffraction kernels, \ie,~point spread function (PSF), and two paired datasets, \ie,~transparent-organic LED (T-OLED) and pentile-organic LED (P-OLED), by mounting a display on top of a traditional digital camera lens. However, they are unaligned with the real UDC degradation due to the lack of high dynamic range (HDR). Based on the measured PSF in real devices, DSICNet~\cite{feng2021removing} generates a larger benchmark with the proposed model-based data synthesis pipeline for ECCV22 challenge~\cite{feng2022mipi}. In addition, since multiple artificial lights at night may introduce different diffraction patterns, nighttime flare removal~\cite{dai2022flare7k} and haze removal~\cite{liu2023nighthazeformer} are also partially similar to the UDC restoration.

Existing works treat UDC image restoration as an inversion problem for the measured PSFs (\ie,~diffraction templates). To eliminate the degradation arising from the PSF, some PSF-related methods~\cite{kwon2021controllable,liu2022udc} use the PSF as a priori to guide the diffraction removal. Further, to avoid the PSF diversity caused by multiple external factors, the PSF-free methods~\cite{panikkasseril2020transform,zhou2021image} directly learn various degradations in UDC images through complex network design. Typically, MSUNet~\cite{zhou2021image}, DAGF~\cite{feng2021removing}, and BNUDC~\cite{koh2022bnudc} propose U-Net~\cite{ronneberger2015u} framework, deep atrous guided filter, and dual-stream framework for UDC restoration, respectively. Recently, AlignFormer~\cite{feng2023generating} proposes the first reference-based framework for non-aligned UDC image restoration.

\subsubsection{Video Restoration.}
Video restoration aims to recover a high-quality video from a low-quality counterpart. Existing video restoration methods can be categorized into two kinds of paradigms: based on sliding-window structure~\cite{kim2018spatio,li2021arvo,wang2019edvr,li2023simple} and based on recurrent structure~\cite{huang2017video,isobe2020video,tao2018scale,sajjadi2018frame}. The methods based on sliding-window structure use adjacent frames within a sliding window as inputs to recover the high-quality frame (\eg,~5 or 7 frames). They mainly use Transformer~\cite{liang2022vrt} or deformable convolutions~\cite{wang2019edvr} to design advanced alignment modules and fuse useful features from adjacent frames. Nevertheless, multi-frame inputs often require higher computational costs, especially when using larger window sizes to model frames at more distance. Rather than only aggregating information from adjacent frames, methods based on the recurrent structure can deliver the relevant features from past frames over time. These methods either devote their attention to designing advanced propagation methods for utilizing frames at longer distances~\cite{chan2022basicvsrpp,liu2022learning}, or exploit powerful attention mechanisms to enhance feature extraction in the recurrent framework~\cite{liang2022recurrent,zhong2023real}. 

However, in contrast to degradation in super-resolution, deblurring, and other low-level tasks, variations in the intensity of incident light cause different degrees of degradation in UDC video. We propose a more promising solution to eliminate various degradations in UDC video restoration by taking full advantage of the recurrent framework.

\section{Problem Formulation and Dataset}

\subsubsection{Problem Formulation.}
Inspired by the real-world imaging process of commercial UDC systems, we follow the existing UDC image restoration works~\cite{koh2022bnudc,zhou2021image} to define the UDC video restoration as the diffraction removal problem. In UDC video, the degradation model of the $t^{\text{th}}$ frame can be formulated as:
\begin{equation}
\label{e1}
    I_{t}=f(\gamma \cdot I_{t}^{GT} \ast k_{t} + n),
\end{equation}
where $I_{t}^{GT}$ and $I_{t}$ denote the clean and degraded frame, respectively. $k_{t}$ is the diffraction kernel (\ie,~PSF), which is the primary factor affecting the visual quality. $\gamma$ and $n$ are the intensity scaling factor and additive noise, respectively.  $\ast$ denotes the convolution operator. $f(\cdot)$ denotes the clamp function used to simulate the channel saturation. Here we omit the non-linear mapping for brevity. 

From~\cite{zhou2021image}, PSF is determined by the screen pattern $p(x,y)$ in the view of the light source. Different from the image restoration that keeps the PSF constant, when the light source changed during video shooting, PSF changes accordingly~\cite{kwon2021controllable}.
As illustrated by Fig.~\ref{fig:image}(b), we follow existing works~\cite{babbar2022homography,ye2021motion,liu2022unsupervised} to simulate the dynamic changes of $k_{t}$ during the motion by computing the inter-frame Homography matrix $H_{t-1\to t}$, formulated as:
\begin{equation}
\label{e2}
    \begin{aligned}
    k_{t} &= \mathcal{T}(k_{t-1},H_{t-1\to t})\\
     &= \left | \mathcal{F}(H^{-1}_{t-1\to t}(\mathcal{F}^{-1}(sqrt(k_{t-1})))) \right | ^{2}, \\
    H&_{t-1\to t} =\mathcal{M}(I_{t-1}^{GT},I_{t}^{GT}),
    \end{aligned}
\end{equation}
where $\mathcal{F}(\cdot)$ and $\mathcal{F}^{-1}(\cdot)$ are the Fourier transform and its inverse transform, respectively. $H^{-1}_{t-1\to t}$ is the inverse matrix of $H_{t-1\to t}$. $\mathcal{T}(\cdot)$ is the transform function that uses the $H^{-1}_{t-1\to t}$ to perspective warp the PSF of the previous frame $k_{t-1}$. $\mathcal{M}(\cdot)$ is the matching part used to compute the Homography matrix between frames.

\subsubsection{Simulated Data.}
To keep the high dynamic range and high resolution of UDC video, we collected a total of 677 HDR videos from YouTube covering various scenarios present in HDRi Haven (\eg,~Outdoor, Skies, Urban, Night, Nature, and so on) and measured the PSF using a commercial ZTE Axon 20 device. Each video consists of 50 frames with a resolution of $1080\times 1920$. 
For each video, we simulated the corresponding degraded video using Eq.~(\ref{e1}), where the PSF $k_{t}$ is dynamically changed by Eq.~(\ref{e2}). To simulate the exhibit of structured flares near strong light sources, brightness augmentation is also applied in each frame. Finally, 627 videos are selected for training, and the remaining 50 videos are for testing randomly.

\subsubsection{Real Data.}
To verify the effectiveness of the D$^2$RNet in real world, we captured 10 raw videos of different scenarios using the same ZTE Axon 20. We keep a high dynamic range and the same resolution with the simulated data.

\section{Methodology}

\begin{figure*}[!tbp]
  \begin{center}
  \includegraphics[width=0.95\linewidth,page=1]{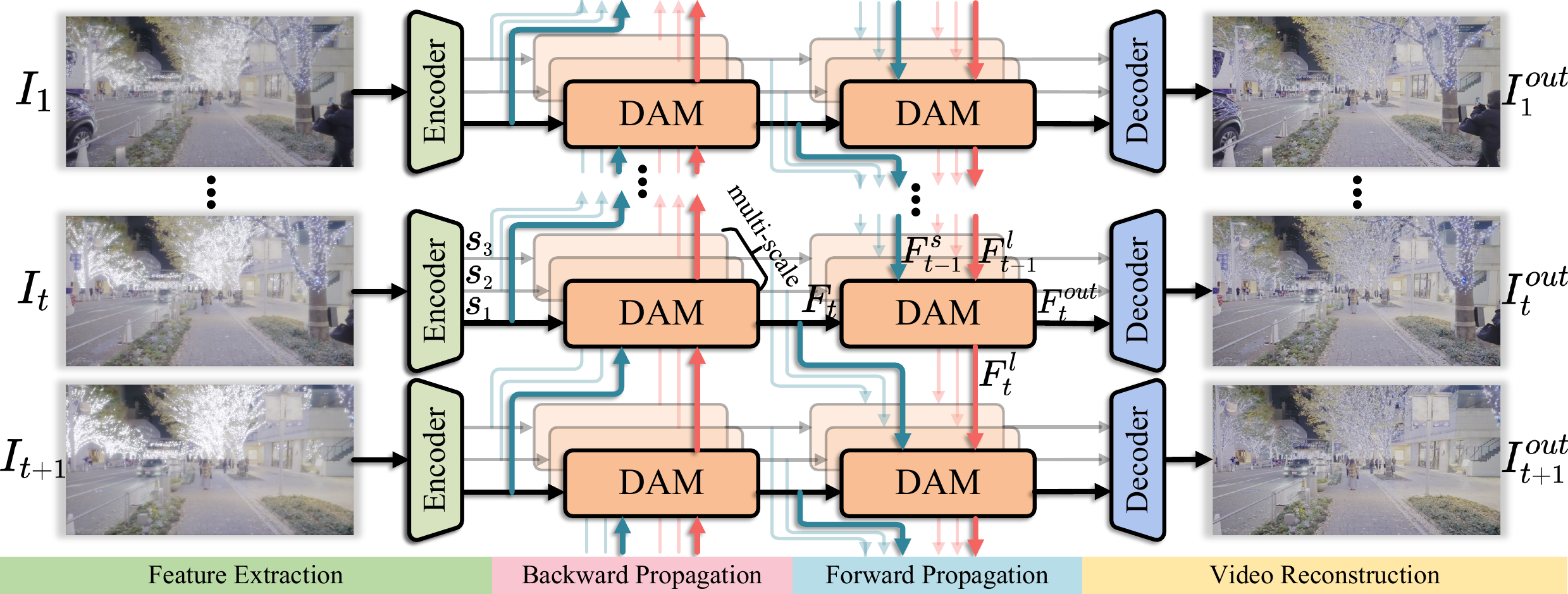}
  \end{center}
  \caption{Overview of D$^2$RNet. It adopts a multi-scale bilateral recurrent architecture. 
  Where an encoder and decoder are used to extract frame features and reconstruct the output frame, respectively. The proposed decoupling attention modules (DAM, see details in Fig.~\ref{fig:dam}) is used to refine the features in both backward and forward propagation, which is supervised at multi-scale.}
  \label{fig:overview}
\end{figure*}

\subsection{Overview of D$^2$RNet}
By analyzing the properties of different degradations (\ie,~flare and haze) caused by variations in incident light intensity, we introduce valuable insight into handling UDC video restoration by decoupling the degradations. 
The overall structure of the proposed D$^2$RNet is shown in Fig.~\ref{fig:overview}. Specifically, given a degraded low-quality sequence $\mathbf{I}_{LQ}=\{I_{t}\in \mathbb{R}^{C\times H\times W}, t \in \{1,2,\dots,T\}\}$, the goal is to recover a high-quality version $\mathbf{I}_{HQ}=\{I_{t}^{out}\in \mathbb{R}^{C\times H\times W}, t \in \{1,2,\dots,T\}\}$. Where $T$, $C$,$H$, and $W$ are the sequence length, channel, height, and width, respectively. The whole model adopts the recurrent architecture that combines multi-scale feature learning and bi-directional propagation. In which, the core decoupling attention module (DAM) refines features during backward and forward propagation.

\subsection{Multi-scale Bi-directional Recurrent Architecture}
\label{mbra}
Inspired by the success of bi-directional recurrent~\cite{chan2022basicvsrpp,huang2015bidirectional} and multi-scale fusion~\cite{cho2021rethinking,zamir2021multi} in low-level tasks, we combine them to enhance video representations. As shown in Fig.~\ref{fig:overview}, from left to right, there is an encoder for extracting multi-scale features, backward and forward propagation for features learning, and a decoder for reconstructing the output frames, respectively.

Formally, take the restoration process of input $I_{t}$ of the $t^{\text{th}}$ frame as an example. First, during feature extraction, we use the contracting path (up-to-down) of UNet~\cite{ronneberger2015u} as the structure of the encoder. This structure tailored for image restoration is broadly considered to enhance local details at different scales~\cite{koh2022bnudc,zamir2021multi}. Specifically, we denote the encoder as $\mathrm{E}(\cdot)$, the output can be obtained by:
\begin{equation}
    F_{t}^{s_1},F_{t}^{s_2}, F_{t}^{s_3}  = \mathrm{E}(I_{t}),
\end{equation}
where $F_{t}^{s_1}\in \mathbb{R}^{C_{s_1}\times \frac{H}{s_1}\times \frac{W}{s_1}}$, $F_{t}^{s_2}\in \mathbb{R}^{C_{s_2}\times \frac{H}{s_2}\times \frac{W}{s_2}}$, and $F_{t}^{s_3}\in \mathbb{R}^{C_{s_3}\times \frac{H}{s_3}\times \frac{W}{s_3}}$ indicate the obtained multi-scale features. 
$C_{s_1}$, $C_{s_2}$, and $C_{s_3}$ are the number of feature channels at scales ${s_1}$, ${s_2}$, and ${s_3}$, respectively, which increases progressively as the spatial resolution of the feature decreases. 

Then, during bi-directional propagation, the proposed DAM, denoted as $\mathrm{DAM}(\cdot)$, refines the features recurrently by inputting the current and the historical features. Take one of the forward propagation as an example (omitting the scale symbols for brevity). This process can be formulated as:
\begin{equation}
    F_{t}^{out},F_{t}^{l}  = \mathrm{DAM}(F_{t},F_{t-1}^{s},F_{t-1}^{l}),
\end{equation}
where $F_{t-1}^{s}=F_{t-1}$ denotes the short-term features from the previous frame, and $F_{t-1}^{l}$ indicates the long-term features from the DAM output of the previous frame. Likewise, DAM is also used for backward propagation and multi-scale.
Besides, to enable better learning of features, we multiplex the features by taking the output of backward propagation as the input of forward propagation.

Finally, we use the expanding path (down to up) of UNet~\cite{ronneberger2015u} as the structure of the decoder, denoted as $\mathrm{D}(\cdot)$. The features after propagation are used to reconstruct the output, formulated as:
\begin{equation}
    I_{t}^{out} = \mathrm{D}(F_{t}^{out,s_1},F_{t}^{out,s_2}, F_{t}^{out,s_3}),
\end{equation}
where $I_{t}^{out}$ is the outut frame. The inputs $F_{t}^{out,s_1}$, $F_{t}^{out,s_2}$, and $F_{t}^{out,s_3}$ indicate multi-scale features output from the bi-directional recurrent propagation.

\subsection{Decoupling Attention Module}
\label{dam}
\begin{figure}[t]
  \begin{center}
  \includegraphics[width=1.0\linewidth,page=1]{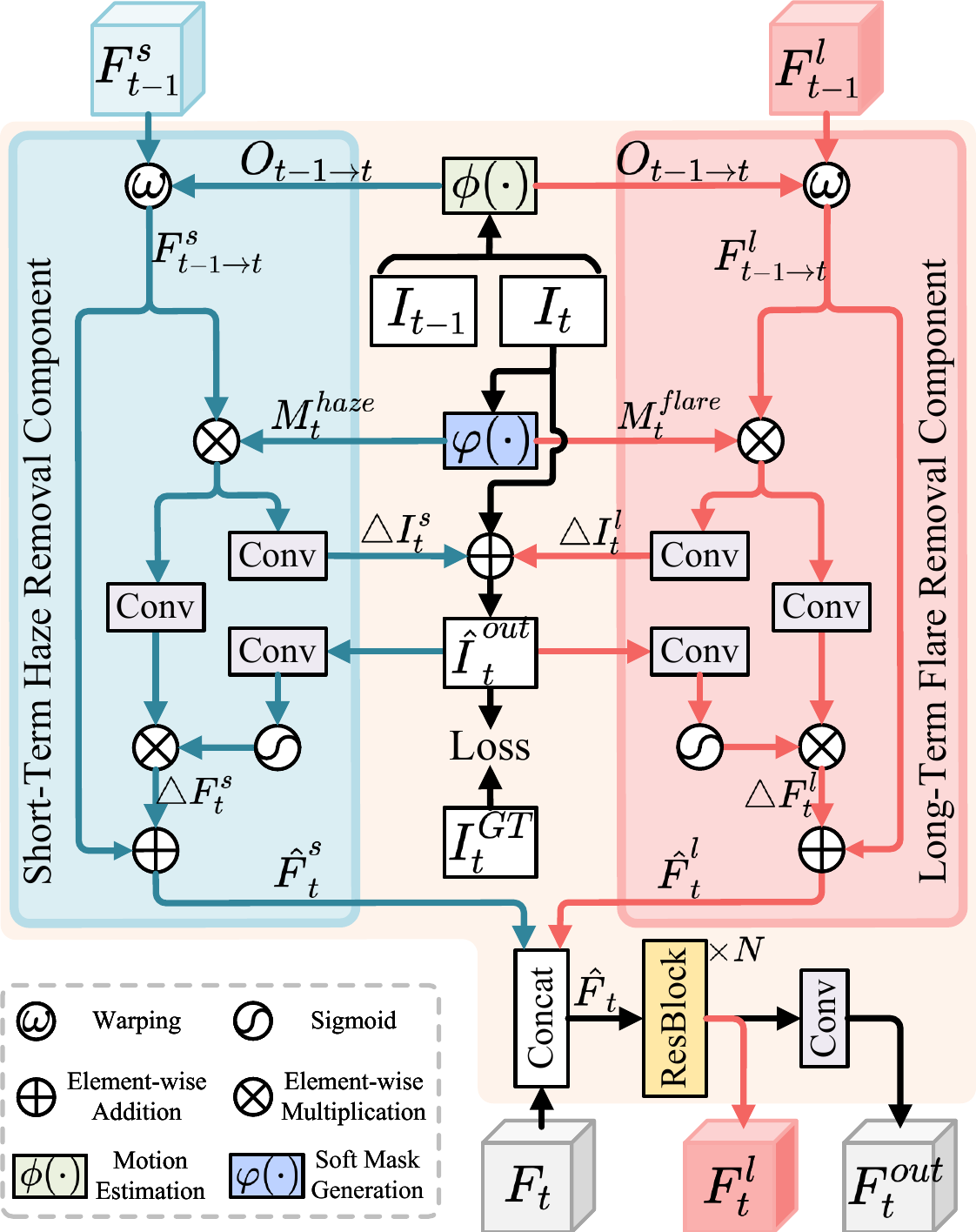}
  \end{center}
  \caption{Structure of the Decoupling Attention Module (DAM). From top to bottom, it mainly consists of a soft mask generation function $\varphi(\cdot)$ for decoupling the flare $M^{flare}_t$ and haze $M^{haze}_t$, and flare and haze removal components handle the respective degradations using long- and short-term features, respectively.}
  \label{fig:dam}
\end{figure}

Along with the diffraction changes with video motion, the distant frame has some complementary features for recovering the lost content due to the flare. Conversely, the removal of haze in the current frame is less correlated with the content in the distant frames, and utilizing more similar features in the spatio-temporal neighborhood is more cost-effective for removing content haze. Therefore, we propose a tailored decoupling attention module (DAM) in Fig.~\ref{fig:dam}.
It includes a soft mask generation function to decouple the flare and haze, and flare and haze removal components handle the respective degradations. We omit scale symbols for brevity.

Formally, we first align the given long-term feature $F_{t-1}^{l}$ and short-term feature $F_{t-1}^{s}$ to the current frame with motion estimation network $\phi(\cdot)$ and warping operation $\mathcal{W}(\cdot)$, formulated as:
\begin{equation}
    \begin{aligned}
    F_{t-1\to t}^{l} &= \mathcal{W}(F^{l}_{t-1},O_{t-1\to t}),\\
    F_{t-1\to t}^{s} &= \mathcal{W}(F^{s}_{t-1},O_{t-1\to t}),\\
    O_{t-1\to t} &= \phi(I_{t-1},I_{t}),
    \end{aligned}
\end{equation}
where $F_{t-1\to t}^{l}$ and $F_{t-1\to t}^{s}$ are the output aligned features. $O_{t-1\to t}$ represents the optical flow. 
Flares are usually caused by strong glare signals and occur with channel saturation. Therefore, inspired by the separation of overexposed regions in HDR images~\cite{cao2023decoupled,eilertsen2017hdr,liu20224d}, an essential soft mask generation function $\varphi(\cdot)$ is proposed to generate the corresponding map of flare $M^{flare}_t$ and haze $M^{haze}_t$, formulated as:
\begin{equation}
    \begin{aligned}
    &M^{flare}_t = \varphi(I_{t}),\ \ \ \ M^{haze}_t = 1- \varphi(I_{t}),\\
    \varphi(&I_{t}^{\{r,g,b\}}) = \frac{max(0,max_{c}(I_{t}^{r},I_{t}^{g},I_{t}^{b})-\tau)}{1-\tau},
    \end{aligned}
\end{equation}
where $max_{c}(\cdot)$ denotes taking the maximum value in the channel dimension. $\tau$ is an empirical parameter used to partition the flare and haze maps. 
$M^{flare}_t$ measures the reliability of the flare and locates the region where the flare occurs.
The value in $M^{flare}_t$ is a linear ramp starting from pixel values at a threshold $\tau$, and ending at the maximum pixel value. $M^{haze}_t$ opposite to it.

Further, benefiting from the progress of the supervised attention mechanism ~\cite{cho2021rethinking,zamir2021multi} in image restoration tasks, we generate intermediate results for supervising the training in each stage. Specifically, with the guidance of $M^{flare}_t$ and $M^{haze}_t$, the generated intermediate result $\hat{I}_t^{out}$ can be formulated as:
\begin{equation}
    \begin{aligned}
    \hat{I}_t^{out} &= I_{t} + \triangle I_{t}^{l} + \triangle I_{t}^{s},\\
    \triangle I_{t}^{l} &= Conv(M_{t}^{flare}\otimes F_{t-1\to t}^{l}),\\
    \triangle I_{t}^{s} &= Conv(M_{t}^{haze}\otimes F_{t-1\to t}^{s}),
    \end{aligned}
\end{equation}
where $Conv(\cdot)$ denotes the convolutional layer. $\otimes$ denotes the element-wise multiplication. Compared to existing restoration methods~\cite{suin2020spatially,zhang2019deep} that directly predict images at each stage and input them to subsequent stages, the introduction of supervision between the intermediate results $\hat{I}_t^{out}$ and the corresponding ground truth in each stage contributes to feature learning and performance gains. 

Then, with the help of supervised $\hat{I}_t^{out}$, we generate attention maps that allow us to preserve the useful features to refine the long-term features $F_{t-1}^{l}$ and short-term features $F_{t-1}^{l}$, formulated as:
\begin{equation}
    \begin{aligned}
    \hat{F}_t^{l} &= F_{t-1\to t}^{l}\oplus  \triangle F_{t}^{l},\ \ \ \ \      \hat{F}_t^{s} = F_{t-1\to t}^{s} \oplus  \triangle F_{t}^{s},\\
    \triangle F_{t}^{l} &= (S(Conv(\hat{I}_t^{out}))) \otimes Conv(M_{t}^{flare}\otimes F_{t-1\to t}^{l}), \\
    \triangle F_{t}^{s} &= (S(Conv(\hat{I}_t^{out}))) \otimes Conv(M_{t}^{haze}\otimes F_{t-1\to t}^{s}),
    \end{aligned}
\end{equation}
where $\hat{F}_t^{l}$ and $\hat{F}_t^{s}$ are the output refined features. $S(\cdot)$ is a sigmoid function for generating attention maps. $\oplus$ denotes the element-wise addition.

Finally, the refined features $\hat{F}_t^{l},\hat{F}_t^{s}$ and the current features $F_{t}$ are concatenated to update the long-term features while outputting features for reconstruction, formulated as:
\begin{equation}
    \begin{aligned}
    &F^{out}_t = Conv(F_{t}^{l}),\\
    F_{t}^{l}&= RBs(C(F_{t},\hat{F}_{t}^{l},\hat{F}_{t}^{s})),
    \end{aligned}
\end{equation}
where $F^{out}_t$ is the output of DAM for the final reconstruction. $F_{t}^{l}$ is the updated long-term feature used for the next frame inference. $RBs(\cdot)$ denotes the $N$ stacked residual blocks. $C(\cdot)$ is feature concatenation along the channel.

\begin{figure*}[ht]
  \begin{center}
  \includegraphics[width=0.90\linewidth,page=1]{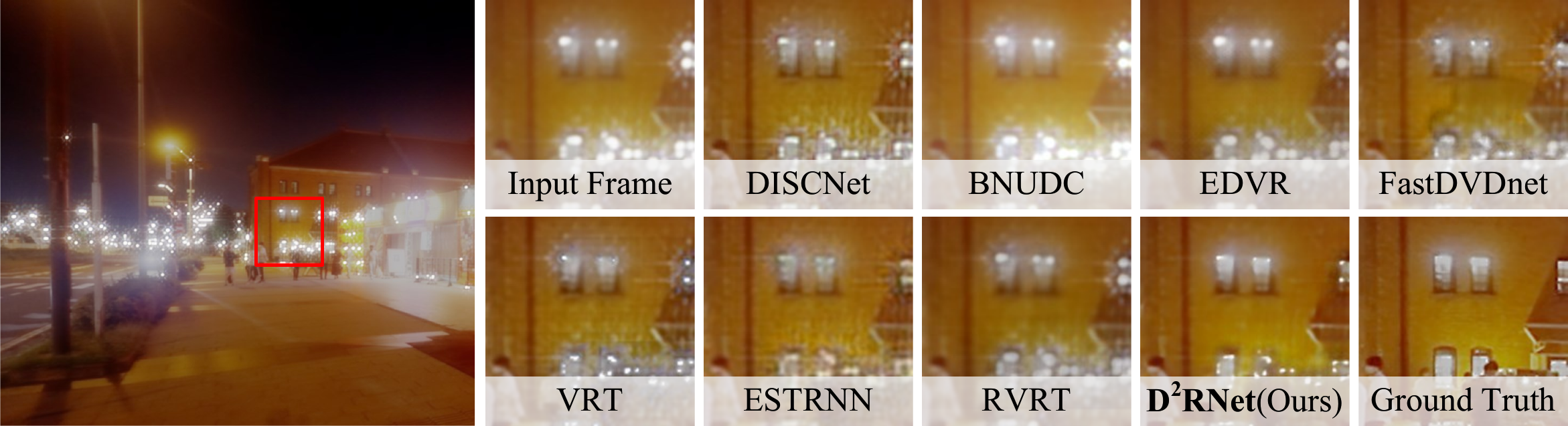}
  \end{center}
  \caption{Visual results on proposed VidUDC33K. The method is shown on the bottom. Zoom in to see better visualization.}
  \label{fig:case}
\end{figure*}

\section{Experiments}

\subsection{Dataset and Metrics}
Since no other datasets are available to study this problem, we compare our D$^2$RNet with other SOTA methods on the proposed VidUDC33K dataset. 
We keep the same evaluation metrics: 1) peak signal-to-noise ratio (PSNR), 2) structural similarity index (SSIM)~\cite{wang2004image} and  3) learned perceptual image patch similarity (LPIPS)~\cite{zhang2018unreasonable} as previous works~\cite{liang2022vrt,liang2022recurrent}.

\begin{table}[t]\small
\centering
       \setlength{\tabcolsep}{2mm}{
       \begin{tabular}{l|c|c|ccc}
        \hline
        
        \hline
        Method     & RT(s)     & \#P(M)  & PSNR$\uparrow$ & SSIM$\uparrow$ & LPIPS$\downarrow$\\ % DISTS
        \hline
        DISCNet     &  0.73   & 3.80   &  28.89  & 0.8405  &  0.2432   \\  %0.1634
        BNUDC       &  0.09  & 4.60 &  28.59 & 0.8398  & 0.2728    \\   % 0.1895
        UDC-UNet       &  0.37  & 5.70 &  28.37 & 0.8361  & 0.2561    \\  
        Alignformer       &  -  & - &  28.96 & 0.8610  & 0.2200    \\  
        EDVR        &   1.17 &  23.6  & 28.71  &  0.8531 &  0.2416   \\
        FastDVDnet  &  0.45 &  2.48  &  28.95 &  0.8638 &  0.2203   \\
        ESTRNN      & 0.20  &   2.47 &  29.54 & 0.8744  & 0.2170    \\
        VRT         &   2.18 &  17.5 &  30.61  & 0.9235  &   0.1397  \\
        RVRT        &   1.68 &  13.9  &  30.89  & 0.9261  &   0.1314 \\% 0.1847
        \textbf{D$^2$RNet}       &  0.44  &  5.76  &  \textbf{31.91} &  \textbf{0.9313} &  \textbf{0.1306}  \\  %0.0878
        \hline
        
        \hline
      \end{tabular}}
        \caption{Quantitative comparison (PSNR(dB)$\uparrow$, SSIM$\uparrow$, and LPIPS$\downarrow$) on the VidUDC33K dataset. RT and \#P indicate the runtimes and parameters, respectively.}\label{tab:results}
\end{table}

\subsection{Training Details}

For fair comparisons, we follow existing works~\cite{liang2022recurrent} to use the pre-trained SPyNet~\cite{ranjan2017optical} for motion estimation. In multi-scale architecture, $s_1$, $s_2$, and $s_3$ correspond to $2\times$, $4\times$, and $8\times$ down-sampling, where the channels of features are 48, 60, and 72, respectively. In DAM, the threshold $\tau$ in the SMG is set to 0.9, and the number of ResBlocks $N$ is set to 5. During training, we use Cosine Annealing scheme and Adam optimizer with $\beta_{1}=0.9$ and $\beta_{2}=0.99$. The learning rates of the motion estimation and other parts are set as $1.25\times 10^{-5}$ and $1\times 10^{-4}$, respectively. We set the batch size as $8$ and the input patch size as $256\times 256$. 
To keep fair comparisons, we augment the data with random horizontal flips, vertical flips, and $90^{\circ}$ rotations. Besides, to enable long-range sequence capability, we use sequences with a length of 30 as inputs. The Charbonnier penalty loss, defined as $\mathcal{L}(x,y)=\sqrt{\|x-y\|^{2}+\varepsilon^2}$ where $\epsilon=10^{-3}$, is applied not only to the whole frames between the restored frame $I^{out}$ and ground truth, but also to the whole frames between the intermediate result $\hat{I}^{out}$ and the ground truth. To stabilize the training, we fix the motion estimation network in the first 5K iterations, and make it trainable later. The total number of iterations is 400K. 

\subsection{Comparisons with State-of-the-art Methods}

We compare our D$^2$RNet with four UDC image restoration models~\cite{feng2021removing,koh2022bnudc,feng2023generating,liu2022udc} and five video restoration models~\cite{liang2022vrt,liang2022recurrent,tassano2020fastdvdnet,wang2019edvr,zhong2023real}. For fair comparisons, we reproduce results with recommended configurations by the authors' officially released codes.

\begin{table}[t] 
  \centering
    \setlength{\tabcolsep}{1.3mm}{
        \begin{tabular}{ccccc|ccc}
            \hline
            
            \hline
             Base    & IS & SHR     & LFR  & SMG & PSNR & SSIM &LPIPS\\
            \hline
            \checkmark     &    &   &   &                    & 30.49  &  0.9220   & 0.1394  \\
            \checkmark     & \checkmark &   &   &             &  31.01 &  0.9277   & 0.1352  \\
            \checkmark     & \checkmark & \checkmark  &   &          &    31.25  &  0.9301   & 0.1344  \\
            \checkmark     & \checkmark &   &  \checkmark &          &    31.37  &  0.9308   & 0.1340  \\
            \checkmark     & \checkmark &  \checkmark &  \checkmark &   &  31.74 &  0.9312   & 0.1314  \\
            \checkmark     & \checkmark &  \checkmark &  \checkmark & \checkmark  &  \textbf{31.91} &  \textbf{0.9313}   &  \textbf{0.1306} \\
            \hline
            
            \hline
        \end{tabular}}
  \caption{Ablation study of each components on the proposed VidUDC33K dataset.}
    \label{fig:by:table} 
\end{table}

\paragraph{Quantitative comparison.}
The performance comparisons on our proposed VidUDC33K dataset are shown in Tab.~\ref{tab:results}. The image-based UDC restoration method (\eg,~Alignformer~\cite{feng2023generating}) cannot exploit temporal information, resulting in poor performance, despite having fewer parameters and runtimes. Methods dedicated to video denoise (\ie,~FastDVDnet~\cite{tassano2020fastdvdnet}) and video blurring (\ie,~ESTRNN~\cite{zhong2023real}) do not yield the ideal performance due to the lack of design to handle diffraction. Moreover, compared to the latest video restoration algorithm (\eg,~RVRT~\cite{liang2022recurrent}), which treat all degradations as equivalent, our method outperforms the latest methods in both objective evaluation metrics PSNR, SSIM and perceptual metrics LPIPS with less runtime and parameters. In particular, our method exceeds the RVRT~\cite{liang2022recurrent} by \textbf{1.02~dB} in PSNR, benefiting from our multi-scale bi-directional recurrent architecture and the design of the decoupled degradation. This large margin demonstrates the power of D$^2$RNet.

\paragraph{Qualitative comparison.}
To further compare the visual qualities of different algorithms, we show visual results restored by our D$^2$RNet and other SOTA methods in Fig.~\ref{fig:case}. It can be observed that compared to other algorithms, D$^2$RNet can simultaneously remove flare at brighter windows and haze elsewhere. It verify that D$^2$RNet has a stronger UDC video restoration capability and has a great improvement in visual quality, especially for flare-rich videos.

\subsection{Ablation Study}
In this section, we first conduct ablation for each component in DAM. After that, we study the effect of the $\tau$ in the SMG and the effect of the multi-scale architecture.

\begin{figure}[t]
    \centering
    \includegraphics[width=1.0\linewidth,page=1]{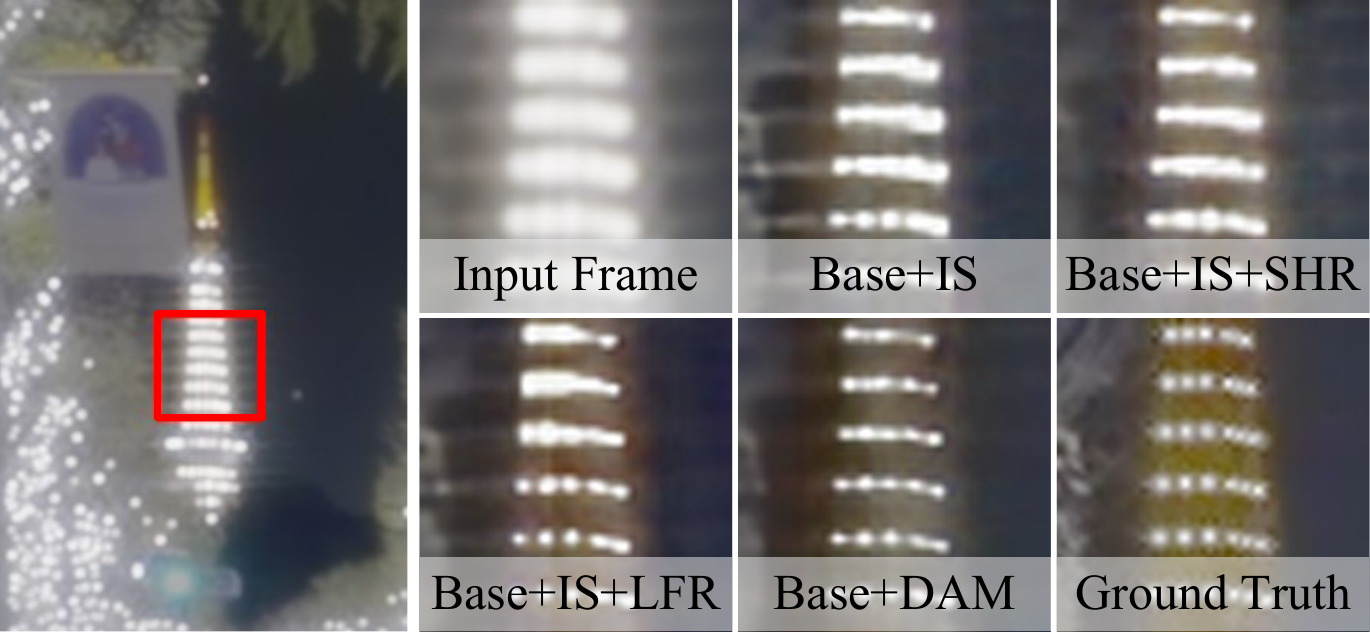}
    \caption{Visual comparison of different components used.} 
    \label{table:by:fig} 
\end{figure}

\begin{table}[t]\small
  \centering
  \begin{tabular}{c|cccc}
    \hline
    
    \hline
    $\tau$     & 0.98  & 0.95  & 0.90 & 0.85 \\
    \hline
    PSNR     &  31.30 &  31.74 & \textbf{31.91}  & 31.78  \\
    SSIM     &  0.9299 & 0.9310  &  \textbf{0.9313} & 0.9310  \\
    \hline
    
    \hline
  \end{tabular}
  \caption{Ablation of $\tau$ in soft mask generation function.}  \label{tab:a1}
\end{table}

\paragraph{Individual components.}
Based on our proposed model, we directly use ResBlock~\cite{he2016deep} to replace the decoupling attention module as the ``Base'' model and progressively add the intermediate supervision (IS), the short-term haze removal component (SHR), the long-term flare removal component (LFR), and the soft mask generation function (SMG) for comparisons. 
As shown in Tab.~\ref{fig:by:table}, the PSNR can be improved from 30.49~dB to 31.74~dB with the addition of IS, SHR, and LFR, verifying the powerful ability of the supervised attention mechanism and the haze/flare removal component. When SMG is involved, degraded haze and flare are decoupled and learned separately, and the performance is improved to 31.91~dB. 
These demonstrate the superiority of each part in DAM. 
We further explore the visual differences as shown in Fig.~\ref{table:by:fig}. LFR can eliminate the content loss caused by flare, and SHR can eliminate the content fuzziness caused by haze. Decoupling flare and haze in UDC videos to remove them separately can produce clearer textures.

\begin{figure}[t]
  \centering
  \includegraphics[width=1.0\linewidth,page=1]{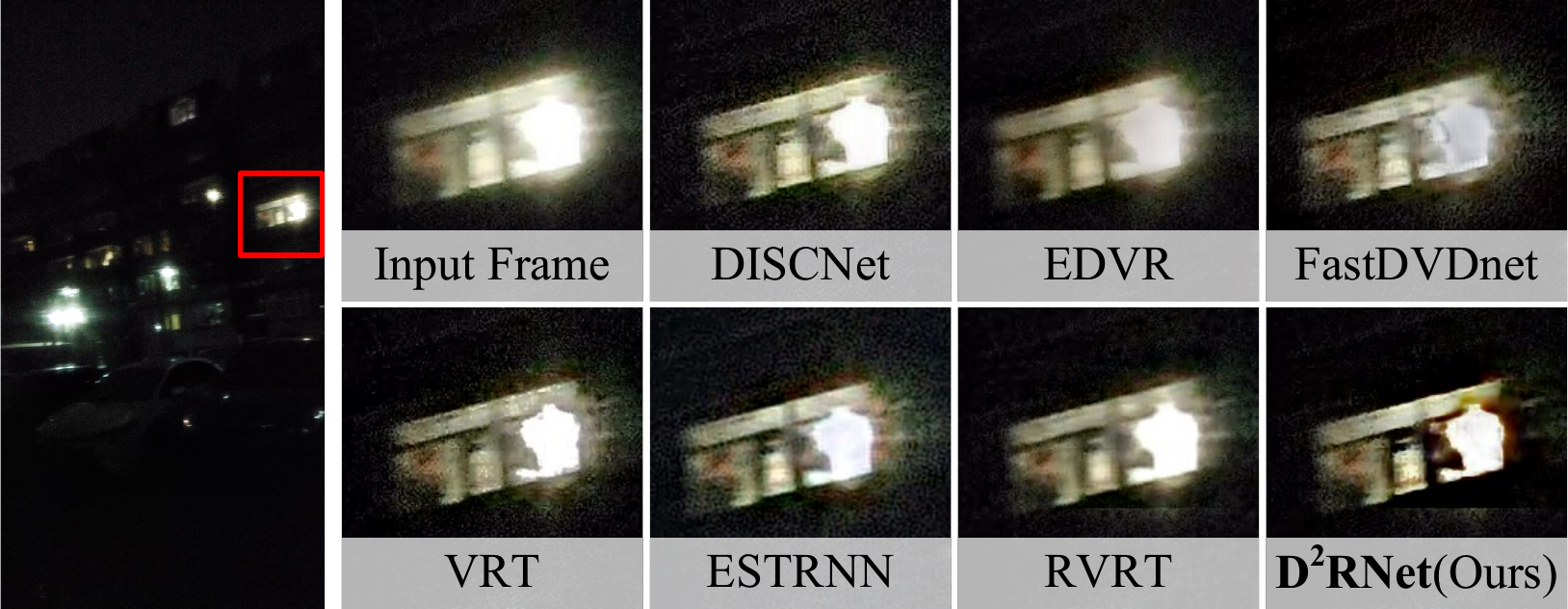}
  \caption{Visual comparison of real UDC video.}
  \label{fig:realcase}
\end{figure}

\begin{table}[t]\small
  \centering
  \begin{tabular}{c|ccc}
    \hline
    
    \hline
    Scale     & (2$\times$)  & (2$\times$,4$\times$)  & (2$\times$,4$\times$,8$\times$) \\
    \hline
    PSNR     &   30.45& 31.36  & \textbf{31.91}  \\
    SSIM     &   0.9192& 0.9308  &  \textbf{0.9313} \\
    \hline
    
    \hline
  \end{tabular}
  \caption{Ablation of the multi-scale structure.}  \label{tab:a2}
\end{table}

\paragraph{The effect of $\tau$.}
To explore the effect of $\tau$ used in the soft mask generation function on performance. In Tab.~\ref{tab:a1}, we use different $\tau$ to decouple the flare and haze. It can be seen that too large $\tau$ does not completely separate the flare region, which affects the recovery of the lost content. On the contrary, too small $\tau$ will result in incomplete removal of content fuzziness. It demonstrates the effectiveness of the soft mask generation function in decoupling video degradation. We set $\tau$ as 0.9 in the final model.

\paragraph{The effect of the multi-scale architecture.}
To alleviate the scale-changing problem in sequences, we discuss the effect of multi-scale architecture on performance. In Tab.~\ref{tab:a2}, we use 2$\times$, 4$\times$, and 8$\times$ to denote $s_1$, $s_2$, and $s_3$ as described in Sec.~\ref{mbra}, respectively. The results show that our method can restore clearer content as the scale increases. The performance can improve PSNR from 30.45~dB to 31.91~dB, indicating that multi-scale architecture can adapt to scale-changing problems in sequences. In our model, we use all three scales to achieve the best performance.

\subsection{Evaluation on Real UDC Video}

In addition to the simulated dataset described above, we conduct compare on real videos collected by the same device. As shown in Fig.~\ref{fig:realcase}, for diffraction-induced flare and haze, our D$^2$RNet can produce clearer textures. This demonstrates the robustness of our D$^2$RNet.

\section{Conclusion}\label{sec:conclusion}

In this paper, we study the effects of the intensity of the incident light and the motion information in UDC video degradation, and introduce a new perspective to handle them by decoupling different types of degradation in advance. 
In particular, we propose a novel video restoration network to enable effective UDC video representation learning, dubbed D$^2$RNet, in which the core decoupling attention module (DAM) provides an effective and targeted solution for eliminating various degradations.
Experimental results show significant performance improvements and clear visual margins between D$^2$RNet and existing SOTA models. To the best of our knowledge, we are the first work to study this problem and propose the first large-scale UDC video benchmark. Our perspective on UDC video has the potential to inspire more diffraction-limited video restoration works. In the future, we will further improve the generality and robustness of our model, and extend it to other low-level vision tasks through more exploration.

\section*{Acknowledgements}
This work was supported in part by the NSFC under Grant 62272380 and 62103317, in part by the Fundamental Research Funds for the Central Universities, China (xzy022023051), in part by the Innovative Leading Talents Scholarship of Xi'an Jiaotong University, in part by the Science and Technology Program of Xi’an, China under Grant 21RGZN0017, in part by the MEGVII Technology.

\bibliography{aaai24}

\begin{thebibliography}{47}
\providecommand{\natexlab}[1]{#1}

\bibitem[{Babbar and Bajaj(2022)}]{babbar2022homography}
Babbar, G.; and Bajaj, R. 2022.
\newblock Homography Theories Used for Image Mapping: A Review.
\newblock In \emph{2022 10th International Conference on Reliability, Infocom Technologies and Optimization (Trends and Future Directions)(ICRITO)}, 1--5. IEEE.

\bibitem[{Cao et~al.(2023)Cao, Zhou, Liu, Wang, and Fan}]{cao2023decoupled}
Cao, G.; Zhou, F.; Liu, K.; Wang, A.; and Fan, L. 2023.
\newblock A decoupled kernel prediction network guided by soft mask for single image {HDR} reconstruction.
\newblock \emph{ACM Transactions on Multimedia Computing, Communications and Applications}, 19(2s): 1--23.

\bibitem[{Chan et~al.(2022)Chan, Zhou, Xu, and Loy}]{chan2022basicvsrpp}
Chan, K.~C.; Zhou, S.; Xu, X.; and Loy, C.~C. 2022.
\newblock {BasicVSR++}: Improving video super-resolution with enhanced propagation and alignment.
\newblock In \emph{CVPR}, 5972--5981.

\bibitem[{Cho et~al.(2021)Cho, Ji, Hong, Jung, and Ko}]{cho2021rethinking}
Cho, S.-J.; Ji, S.-W.; Hong, J.-P.; Jung, S.-W.; and Ko, S.-J. 2021.
\newblock Rethinking coarse-to-fine approach in single image deblurring.
\newblock In \emph{ICCV}, 4641--4650.

\bibitem[{Dai et~al.(2022)Dai, Li, Zhou, Feng, and Loy}]{dai2022flare7k}
Dai, Y.; Li, C.; Zhou, S.; Feng, R.; and Loy, C.~C. 2022.
\newblock Flare7k: A phenomenological nighttime flare removal dataset.
\newblock \emph{NeurIPS}, 35: 3926--3937.

\bibitem[{Eilertsen et~al.(2017)Eilertsen, Kronander, Denes, Mantiuk, and Unger}]{eilertsen2017hdr}
Eilertsen, G.; Kronander, J.; Denes, G.; Mantiuk, R.~K.; and Unger, J. 2017.
\newblock {HDR} image reconstruction from a single exposure using deep CNNs.
\newblock \emph{ACM TOG}, 36(6): 1--15.

\bibitem[{Feng et~al.(2023)Feng, Li, Chen, Li, Gu, and Loy}]{feng2023generating}
Feng, R.; Li, C.; Chen, H.; Li, S.; Gu, J.; and Loy, C.~C. 2023.
\newblock Generating Aligned Pseudo-Supervision from Non-Aligned Data for Image Restoration in Under-Display Camera.
\newblock In \emph{CVPR}, 5013--5022.

\bibitem[{Feng et~al.(2021)Feng, Li, Chen, Li, Loy, and Gu}]{feng2021removing}
Feng, R.; Li, C.; Chen, H.; Li, S.; Loy, C.~C.; and Gu, J. 2021.
\newblock Removing diffraction image artifacts in under-display camera via dynamic skip connection network.
\newblock In \emph{CVPR}, 662--671.

\bibitem[{Feng et~al.(2022)Feng, Li, Zhou, Sun, Zhu, Jiang, Yang, Loy, and Gu}]{feng2022mipi}
Feng, R.; Li, C.; Zhou, S.; Sun, W.; Zhu, Q.; Jiang, J.; Yang, Q.; Loy, C.~C.; and Gu, J. 2022.
\newblock Mipi 2022 challenge on under-display camera image restoration: Methods and results.
\newblock \emph{arXiv preprint arXiv:2209.07052}.

\bibitem[{He et~al.(2016)He, Zhang, Ren, and Sun}]{he2016deep}
He, K.; Zhang, X.; Ren, S.; and Sun, J. 2016.
\newblock Deep residual learning for image recognition.
\newblock In \emph{CVPR}, 770--778.

\bibitem[{Huang, Wang, and Wang(2015)}]{huang2015bidirectional}
Huang, Y.; Wang, W.; and Wang, L. 2015.
\newblock Bidirectional recurrent convolutional networks for multi-frame super-resolution.
\newblock \emph{NeurIPS}, 28.

\bibitem[{Huang, Wang, and Wang(2017)}]{huang2017video}
Huang, Y.; Wang, W.; and Wang, L. 2017.
\newblock Video super-resolution via bidirectional recurrent convolutional networks.
\newblock \emph{IEEE TPAMI}, 40(4): 1015--1028.

\bibitem[{Isobe et~al.(2020)Isobe, Jia, Gu, Li, Wang, and Tian}]{isobe2020video}
Isobe, T.; Jia, X.; Gu, S.; Li, S.; Wang, S.; and Tian, Q. 2020.
\newblock Video super-resolution with recurrent structure-detail network.
\newblock In \emph{ECCV}, 645--660. Springer.

\bibitem[{Kim et~al.(2018)Kim, Sajjadi, Hirsch, and Scholkopf}]{kim2018spatio}
Kim, T.~H.; Sajjadi, M.~S.; Hirsch, M.; and Scholkopf, B. 2018.
\newblock Spatio-temporal transformer network for video restoration.
\newblock In \emph{ECCV}, 106--122.

\bibitem[{Koh, Lee, and Yoon(2022)}]{koh2022bnudc}
Koh, J.; Lee, J.; and Yoon, S. 2022.
\newblock Bnudc: A two-branched deep neural network for restoring images from under-display cameras.
\newblock In \emph{CVPR}, 1950--1959.

\bibitem[{Kwon et~al.(2021)Kwon, Kang, Lee, Lee, Lee, Yoo, and Han}]{kwon2021controllable}
Kwon, K.; Kang, E.; Lee, S.; Lee, S.-J.; Lee, H.-E.; Yoo, B.; and Han, J.-J. 2021.
\newblock Controllable image restoration for under-display camera in smartphones.
\newblock In \emph{CVPR}, 2073--2082.

\bibitem[{Li et~al.(2023)Li, Shi, Zhang, Cheung, See, Wang, Qin, and Li}]{li2023simple}
Li, D.; Shi, X.; Zhang, Y.; Cheung, K.~C.; See, S.; Wang, X.; Qin, H.; and Li, H. 2023.
\newblock A Simple Baseline for Video Restoration With Grouped Spatial-Temporal Shift.
\newblock In \emph{CVPR}, 9822--9832.

\bibitem[{Li et~al.(2021)Li, Xu, Zhang, Yu, Zhong, Ren, Suominen, and Li}]{li2021arvo}
Li, D.; Xu, C.; Zhang, K.; Yu, X.; Zhong, Y.; Ren, W.; Suominen, H.; and Li, H. 2021.
\newblock Arvo: Learning all-range volumetric correspondence for video deblurring.
\newblock In \emph{CVPR}, 7721--7731.

\bibitem[{Liang et~al.(2022{\natexlab{a}})Liang, Cao, Fan, Zhang, Ranjan, Li, Timofte, and Van~Gool}]{liang2022vrt}
Liang, J.; Cao, J.; Fan, Y.; Zhang, K.; Ranjan, R.; Li, Y.; Timofte, R.; and Van~Gool, L. 2022{\natexlab{a}}.
\newblock {VRT}: A video restoration transformer.
\newblock \emph{arXiv preprint arXiv:2201.12288}.

\bibitem[{Liang et~al.(2022{\natexlab{b}})Liang, Fan, Xiang, Ranjan, Ilg, Green, Cao, Zhang, Timofte, and Gool}]{liang2022recurrent}
Liang, J.; Fan, Y.; Xiang, X.; Ranjan, R.; Ilg, E.; Green, S.; Cao, J.; Zhang, K.; Timofte, R.; and Gool, L.~V. 2022{\natexlab{b}}.
\newblock Recurrent video restoration transformer with guided deformable attention.
\newblock \emph{NeurIPS}, 35: 378--393.

\bibitem[{Lin et~al.(2022)Lin, Cai, Hu, Wang, Yan, Zou, Ding, Zhang, Timofte, and Van~Gool}]{lin2022flow}
Lin, J.; Cai, Y.; Hu, X.; Wang, H.; Yan, Y.; Zou, X.; Ding, H.; Zhang, Y.; Timofte, R.; and Van~Gool, L. 2022.
\newblock Flow-guided sparse transformer for video deblurring.
\newblock \emph{arXiv preprint arXiv:2201.01893}.

\bibitem[{Liu et~al.(2023{\natexlab{a}})Liu, Wang, Li, Wang, and Qian}]{liu2023fsi}
Liu, C.; Wang, X.; Li, S.; Wang, Y.; and Qian, X. 2023{\natexlab{a}}.
\newblock FSI: Frequency and Spatial Interactive Learning for Image Restoration in Under-Display Cameras.
\newblock In \emph{ICCV}, 12537--12546.

\bibitem[{Liu et~al.(2022{\natexlab{a}})Liu, Yang, Fu, and Qian}]{liu2022learning}
Liu, C.; Yang, H.; Fu, J.; and Qian, X. 2022{\natexlab{a}}.
\newblock Learning trajectory-aware transformer for video super-resolution.
\newblock In \emph{CVPR}, 5687--5696.

\bibitem[{Liu et~al.(2023{\natexlab{b}})Liu, Yang, Fu, and Qian}]{liu20224d}
Liu, C.; Yang, H.; Fu, J.; and Qian, X. 2023{\natexlab{b}}.
\newblock {4D LUT}: learnable context-aware 4d lookup table for image enhancement.
\newblock \emph{IEEE TIP}, 32: 4742--4756.

\bibitem[{Liu et~al.(2023{\natexlab{c}})Liu, Yang, Fu, and Qian}]{liu2022ttvfi}
Liu, C.; Yang, H.; Fu, J.; and Qian, X. 2023{\natexlab{c}}.
\newblock {TTVFI}: Learning trajectory-aware transformer for video frame interpolation.
\newblock \emph{IEEE TIP}.

\bibitem[{Liu et~al.(2022{\natexlab{b}})Liu, Lu, Jiang, Ye, Wang, and Zeng}]{liu2022unsupervised}
Liu, S.; Lu, Y.; Jiang, H.; Ye, N.; Wang, C.; and Zeng, B. 2022{\natexlab{b}}.
\newblock Unsupervised Global and Local Homography Estimation With Motion Basis Learning.
\newblock \emph{IEEE TPAMI}.

\bibitem[{Liu et~al.(2022{\natexlab{c}})Liu, Hu, Chen, and Dong}]{liu2022udc}
Liu, X.; Hu, J.; Chen, X.; and Dong, C. 2022{\natexlab{c}}.
\newblock UDC-UNet: Under-Display Camera Image Restoration via U-shape Dynamic Network.
\newblock In \emph{ECCV}, 113--129. Springer.

\bibitem[{Liu et~al.(2023{\natexlab{d}})Liu, Yan, Chen, Ye, Ren, and Chen}]{liu2023nighthazeformer}
Liu, Y.; Yan, Z.; Chen, S.; Ye, T.; Ren, W.; and Chen, E. 2023{\natexlab{d}}.
\newblock Nighthazeformer: Single nighttime haze removal using prior query transformer.
\newblock In \emph{ACM MM}, 4119--4128.

\bibitem[{Panikkasseril~Sethumadhavan et~al.(2020)Panikkasseril~Sethumadhavan, Puthussery, Kuriakose, and Charangatt~Victor}]{panikkasseril2020transform}
Panikkasseril~Sethumadhavan, H.; Puthussery, D.; Kuriakose, M.; and Charangatt~Victor, J. 2020.
\newblock Transform domain pyramidal dilated convolution networks for restoration of under display camera images.
\newblock In \emph{ECCVW}, 364--378. Springer.

\bibitem[{Qin et~al.(2021)Qin, Qiu, Li, Yu, and Yang}]{qin2021p}
Qin, Z.; Qiu, R.; Li, M.; Yu, X.; and Yang, B.-R. 2021.
\newblock P-78: Simulator-Based Efficient Panel Design and Image Retrieval for Under-Display Cameras.
\newblock In \emph{SID Symposium Digest of Technical Papers}, volume~52, 1372--1375. Wiley Online Library.

\bibitem[{Ranjan and Black(2017)}]{ranjan2017optical}
Ranjan, A.; and Black, M.~J. 2017.
\newblock Optical flow estimation using a spatial pyramid network.
\newblock In \emph{CVPR}, 4161--4170.

\bibitem[{Ronneberger, Fischer, and Brox(2015)}]{ronneberger2015u}
Ronneberger, O.; Fischer, P.; and Brox, T. 2015.
\newblock U-net: Convolutional networks for biomedical image segmentation.
\newblock In \emph{MICCAI}, 234--241. Springer.

\bibitem[{Sajjadi, Vemulapalli, and Brown(2018)}]{sajjadi2018frame}
Sajjadi, M.~S.; Vemulapalli, R.; and Brown, M. 2018.
\newblock Frame-recurrent video super-resolution.
\newblock In \emph{CVPR}, 6626--6634.

\bibitem[{Suin, Purohit, and Rajagopalan(2020)}]{suin2020spatially}
Suin, M.; Purohit, K.; and Rajagopalan, A. 2020.
\newblock Spatially-attentive patch-hierarchical network for adaptive motion deblurring.
\newblock In \emph{CVPR}, 3606--3615.

\bibitem[{Tao et~al.(2018)Tao, Gao, Shen, Wang, and Jia}]{tao2018scale}
Tao, X.; Gao, H.; Shen, X.; Wang, J.; and Jia, J. 2018.
\newblock Scale-recurrent network for deep image deblurring.
\newblock In \emph{CVPR}, 8174--8182.

\bibitem[{Tassano, Delon, and Veit(2020)}]{tassano2020fastdvdnet}
Tassano, M.; Delon, J.; and Veit, T. 2020.
\newblock Fast{DVD}net: Towards real-time deep video denoising without flow estimation.
\newblock In \emph{CVPR}, 1354--1363.

\bibitem[{Wang et~al.(2019)Wang, Chan, Yu, Dong, and Change~Loy}]{wang2019edvr}
Wang, X.; Chan, K.~C.; Yu, K.; Dong, C.; and Change~Loy, C. 2019.
\newblock {EDVR}: Video restoration with enhanced deformable convolutional networks.
\newblock In \emph{CVPRW}, 0--0.

\bibitem[{Wang et~al.(2022)Wang, Lu, Gao, Wang, Zhong, Zheng, and Yamashita}]{wang2022efficient}
Wang, Y.; Lu, Y.; Gao, Y.; Wang, L.; Zhong, Z.; Zheng, Y.; and Yamashita, A. 2022.
\newblock Efficient video deblurring guided by motion magnitude.
\newblock In \emph{ECCV}, 413--429. Springer.

\bibitem[{Wang et~al.(2004)Wang, Bovik, Sheikh, and Simoncelli}]{wang2004image}
Wang, Z.; Bovik, A.~C.; Sheikh, H.~R.; and Simoncelli, E.~P. 2004.
\newblock Image quality assessment: from error visibility to structural similarity.
\newblock \emph{IEEE TIP}, 13(4): 600--612.

\bibitem[{Ye et~al.(2021)Ye, Wang, Fan, and Liu}]{ye2021motion}
Ye, N.; Wang, C.; Fan, H.; and Liu, S. 2021.
\newblock Motion basis learning for unsupervised deep homography estimation with subspace projection.
\newblock In \emph{ICCV}, 13117--13125.

\bibitem[{Zamir et~al.(2021)Zamir, Arora, Khan, Hayat, Khan, Yang, and Shao}]{zamir2021multi}
Zamir, S.~W.; Arora, A.; Khan, S.; Hayat, M.; Khan, F.~S.; Yang, M.-H.; and Shao, L. 2021.
\newblock Multi-stage progressive image restoration.
\newblock In \emph{CVPR}, 14821--14831.

\bibitem[{Zhang et~al.(2019)Zhang, Dai, Li, and Koniusz}]{zhang2019deep}
Zhang, H.; Dai, Y.; Li, H.; and Koniusz, P. 2019.
\newblock Deep stacked hierarchical multi-patch network for image deblurring.
\newblock In \emph{CVPR}, 5978--5986.

\bibitem[{Zhang, Xie, and Yao(2022)}]{zhang2022spatio}
Zhang, H.; Xie, H.; and Yao, H. 2022.
\newblock Spatio-temporal deformable attention network for video deblurring.
\newblock In \emph{ECCV}, 581--596. Springer.

\bibitem[{Zhang et~al.(2018)Zhang, Isola, Efros, Shechtman, and Wang}]{zhang2018unreasonable}
Zhang, R.; Isola, P.; Efros, A.~A.; Shechtman, E.; and Wang, O. 2018.
\newblock The unreasonable effectiveness of deep features as a perceptual metric.
\newblock In \emph{CVPR}, 586--595.

\bibitem[{Zhong et~al.(2023)Zhong, Gao, Zheng, Zheng, and Sato}]{zhong2023real}
Zhong, Z.; Gao, Y.; Zheng, Y.; Zheng, B.; and Sato, I. 2023.
\newblock Real-world video deblurring: A benchmark dataset and an efficient recurrent neural network.
\newblock \emph{IJCV}, 131(1): 284--301.

\bibitem[{Zhou et~al.(2020)Zhou, Kwan, Tolentino, Emerton, Lim, Large, Fu, Pan, Li, Yang et~al.}]{zhou2020udc}
Zhou, Y.; Kwan, M.; Tolentino, K.; Emerton, N.; Lim, S.; Large, T.; Fu, L.; Pan, Z.; Li, B.; Yang, Q.; et~al. 2020.
\newblock UDC 2020 challenge on image restoration of under-display camera: Methods and results.
\newblock In \emph{ECCVW}, 337--351. Springer.

\bibitem[{Zhou et~al.(2021)Zhou, Ren, Emerton, Lim, and Large}]{zhou2021image}
Zhou, Y.; Ren, D.; Emerton, N.; Lim, S.; and Large, T. 2021.
\newblock Image restoration for under-display camera.
\newblock In \emph{CVPR}, 9179--9188.

\end{thebibliography}

\end{document}